\documentclass[10pt,twocolumn,letterpaper]{article}
\usepackage[pagenumbers]{cvpr}

\usepackage{graphicx}
\usepackage{amsmath}
\usepackage{amssymb}
\usepackage{booktabs}
\usepackage{balance}
\usepackage{lipsum}
\usepackage{caption}
\usepackage{algorithm}
\usepackage{algorithmic}
\usepackage{enumerate}
\usepackage{amsmath,amsthm,amssymb}
\usepackage{engord}
\usepackage{bm}
\usepackage{bbm}
\usepackage{amstext}
\usepackage{comment}
\usepackage{color}
\usepackage[x11names,table]{xcolor}
\usepackage{ctable}
\usepackage{booktabs}
\usepackage{multirow}
\usepackage{mathtools}
\usepackage{setspace}
\usepackage{pifont}
\usepackage{dsfont}
\usepackage[inline]{enumitem}
\usepackage[toc,page]{appendix}
\usepackage[pagebackref=true,breaklinks=true,letterpaper=true,colorlinks,citecolor=citecolor,bookmarks=true]{hyperref}
\usepackage[capitalize]{cleveref}

\usepackage[accsupp]{axessibility}

\definecolor{Gray}{gray}{0.9}
\definecolor{White}{gray}{1}
\DeclareMathAlphabet\mathbfcal{OMS}{cmsy}{b}{n}
\definecolor{DGray}{gray}{0.8}
\definecolor{Gray}{gray}{0.9}
\definecolor{White}{gray}{1}
\definecolor{citecolor}{HTML}{0071bc}
\crefname{section}{Sec.}{Secs.}
\Crefname{section}{Section}{Sections}
\Crefname{table}{Table}{Tables}
\crefname{table}{Tab.}{Tabs.}
\newcommand{\tabincell}[2]{\begin{tabular}{@{}#1@{}}#2\end{tabular}}

\newcommand{\suppl}{\textbf{Appx}\xspace}

\newcommand{\customfootnotetext}[2]{{% Group to localize change to footnote
    \renewcommand{\thefootnote}{#1}% Update footnote counter representation
    \footnotetext[0]{#2} % Print footnote text
}}

\begin{document}

\title{
    ArtiBoost: Boosting Articulated 3D Hand-Object Pose Estimation via \\
    Online Exploration and Synthesis
}

\author{
    {$^{1}$Kailin Li\textsuperscript{$\bm{\star}$},}
    ~{$^{1,2}$Lixin Yang\textsuperscript{$\bm{\star}$},}
    ~{$^{1}$Xinyu Zhan,}
    ~{$^{1}$Jun Lv,}
    ~{$^{1,2}$Wenqiang Xu,}
    ~{$^{1}$Jiefeng Li,}
    ~{$^{1,2}$Cewu Lu\textsuperscript{$\dagger$}}\\
   {$^{1}$Shanghai Jiao Tong University, China}~~~{$^{2}$Shanghai Qi Zhi Institute, China}\\
{\tt\small \{{kailinli}, {siriusyang}, {kelvin34501}, {LyuJune\_SJTU}, {vinjohn},  {ljf\_likit}, {lucewu}\}@{sjtu.edu.cn}}
}

\maketitle

\customfootnotetext{$\bm{\star}$}{ The first two authors contribute equally.}
\customfootnotetext{$\dagger$}{ 
    Cewu Lu is the corresponding author. He is the member of Qing Yuan Research Institute and MoE Key Lab of Artificial Intelligence, AI Institute, Shanghai Jiao Tong University, and Shanghai Qi Zhi Institute, China.
}

% ####### main text ########

\begin{abstract}
    \vspace{-2mm}
    Estimating the articulated 3D hand-object pose from a single RGB image is a highly ambiguous and challenging problem, requiring large-scale datasets that contain diverse hand poses, object types, and camera viewpoints.
    Most real-world datasets lack these diversities. 
    In contrast, data synthesis can easily ensure those diversities separately. 
    However, constructing both valid and diverse hand-object interactions and efficiently learning from the vast synthetic data is still challenging. 
    To address the above issues, we propose ArtiBoost, a lightweight online data enhancement method.
    ArtiBoost can cover diverse hand-object poses and camera viewpoints through sampling in a Composited hand-object Configuration and Viewpoint space (CCV-space) 
    and can adaptively enrich the current hard-discernable items by loss-feedback and sample re-weighting.
    ArtiBoost alternatively performs data exploration and synthesis within a learning pipeline, and those synthetic data are blended into real-world source data for training.
    We apply ArtiBoost on a simple learning baseline network and witness the performance boost on several hand-object benchmarks. 
    Our models and code are available at \url{https://github.com/lixiny/ArtiBoost}.  
\end{abstract}

\vspace{-5mm}\section{Introduction}
\vspace{-1mm}
Articulated bodies, such as the human hand, body, and linkage mechanism, can be observed every day in our life. 
Their joints, links, and movable parts depict the functionality of the articulation body. 
Extracting their transient configuration from image or video sequence, which is often referred to as \textit{Pose Estimation} \cite{li2021hybrik, chen2021handmesh, li2020ancsh}, 
can benefit many downstream tasks in robotics and augment reality. 
Pose estimation for multi-body articulations is especially challenging as it suffers from severe self- or mutual occlusion.
In this work, we paid attention to a certain type of multi-body articulations -- 
composited hand and object poses during their interaction \cite{hasson2019obman, hasson2020leveraging, hasson21homan, yang2021cpf, grady2021contactopt, rhoi2020, liu2021semi, doosti2020hope, hampali2020ho3dv2, huang2020hotnet, karunratanakul2020graspingfield}. 
Hands are the primary means by which humans manipulate objects in the real-world, and the hand-object pose estimation (HOPE) task holds great potential for understanding human behavior \cite{goyal2017somethingsomething, materzynska2020somethingelse, kwon2021h2o, garcia2018first, tekin2019h+o}.

\begin{figure}[!t]
  \begin{center}
      \includegraphics[width=0.95\linewidth]{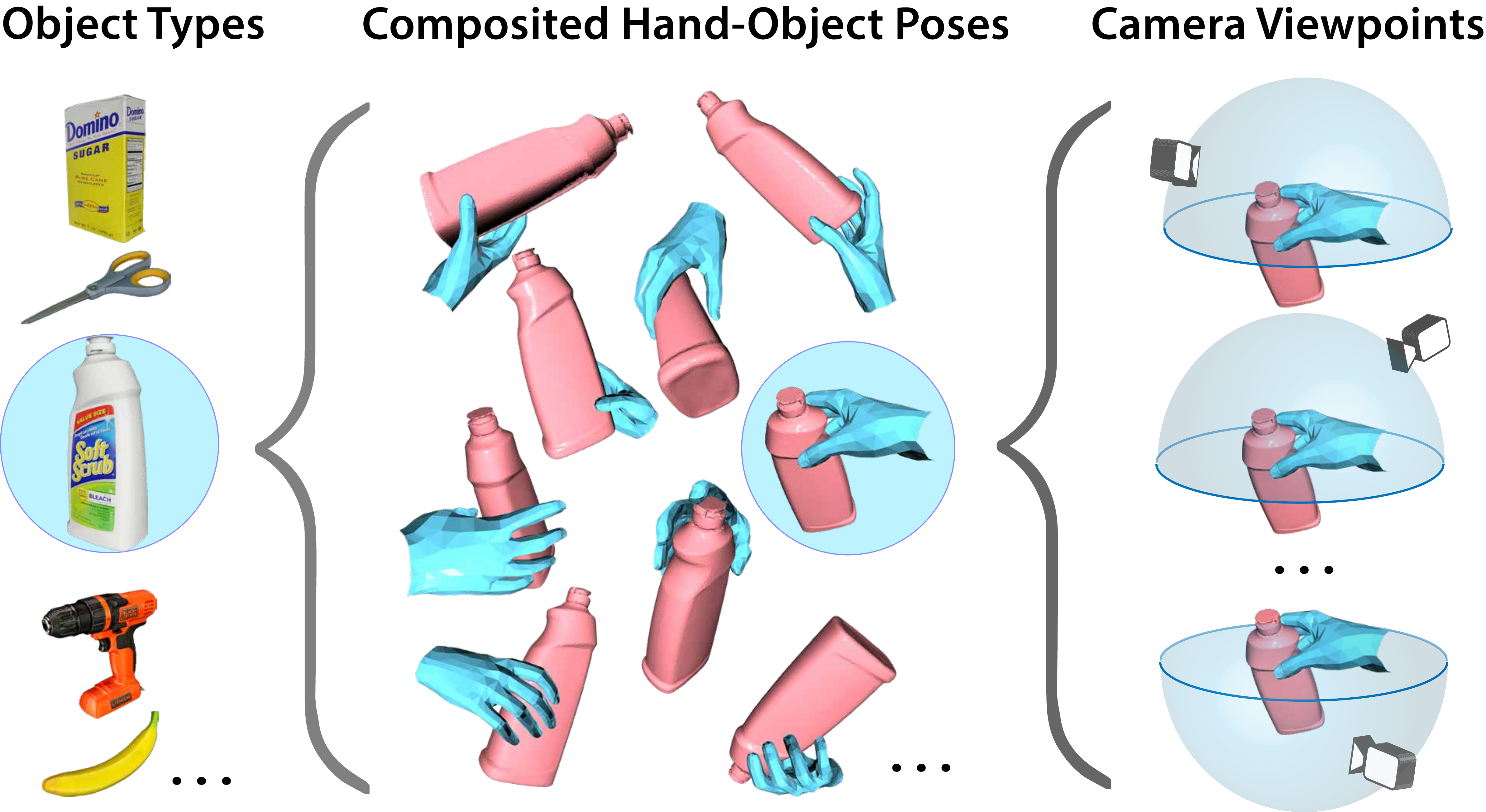}
  \end{center}\vspace{-5mm}
      \caption{\textbf{An intuitive illustration of the CCV-space.}}\vspace{-4.5mm}
  \label{fig:capture_ccvspace}
\end{figure}

As the degrees of freedom (DoF) grows, the proper amount of data to cover the pose distribution has grown exponentially. 
More than the most common articulation bodies, the human hand has 16 joints and approximately 21 DoF. 
Preparing such diverse training data for the HOPE task can be very challenging.  
The real-world recording and annotation methods \cite{brahmbhatt2020contactpose, garcia2018first, hampali2020ho3dv2, chao2021dexycb} tend to hinder the pose diversity. 
For example, the multi-view-based approaches \cite{brahmbhatt2020contactpose, hampali2020ho3dv2} require the subject to maintain a static grasping pose in a video sequence. 
As a result, their recording process is inefficient, and their pose diversity is insufficient.
In contrast, data synthesis is efficient and annotation-free, and has been widely adopted in single-body N-D pose estimation \cite{yan2021ultrapose, varol21_surreact, zimmermann2017rhd, ge20193d, chen2021MobRecon, liu2022towards}. 
However, these methods are ineligible for multi-body articulation, in which the poses are restricted by mutual contact and obstruction. 
Data synthesis for HOPE tasks requires us to simulate virtual hand-object interaction (grasp) that mimics the underlying pose distribution of their real-world counterparts.
Conventional method either manually articulated \cite{romero2010_handsInAction, Rogez2015_GUN71, corona2020ganhand} the hand model for grasp,  
or relegated \cite{hasson2019obman, brahmbhatt2019contactgrasp, kokic2020learningTOG} grasp synthesis to an off-the-shelf grasping simulator: GraspIt \cite{graspit}.
However, the manual methods are difficult to scale their data in large amounts, and the simulation method also sacrificed the diversity of hand poses.~GraspIt optimizes for hand-crafted grasp metrics \cite{ferrari1992_optimalGrasps} that do not reflect the pose distribution of a 21-DoF dexterous hand. 
Besides, even with a vast amount of synthetic grasps, not every hand-object configuration is helpful for training. For example, similar configurations may have already been observed multiple times, and those easily discernable samples may have a frequent appearance. 
Hence, offline data synthesis, without repeatedly communicating with the model during training, is still considered inefficient for a learning task.

To address the above issues, we propose an online data enhancement method \textbf{ArtiBoost}, to effectively \textbf{boost} the \textbf{arti}culated hand-object pose estimation via two alternative steps, namely exploration and synthesis.
First, to describe the observation of hand-object interaction, we design a three-dimensional discrete space: \textbf{C}omposited hand-object \textbf{C}onfiguration and \textbf{V}iewpoint space (CCV-space) where object types, hand pose, and viewpoint are its components.
Second, to construct valid and diverse hand-object poses in CCV-space, we design a fitting-based grasp synthesis method that exploits the contact constraints \cite{yang2021cpf} between hand and object vertices to simulate MANO hand \cite{romero2017embodied} grasping a given object.
After the CCV-space is established, we next describe how ArtiBoost enhances the HOPE tasks.

At the exploration step, ArtiBoost explores the CCV-space and samples different hand-object-viewpoint triplets from it. 
Then at the synthesis step, the hand and object in the triplet will be rendered on the image from the viewpoint in the triplet. 
These synthetic images are mixed with the real-world source images in batches to train the HOPE model. 
Later, the training losses are fed back to the exploration step and guide it to re-weight the current hard-discernable triplets for the next round of sampling.
With such communication in the training loop, ArtiBoost can adaptively adjust its sampling weights to select more hard-discernable data for the current HOPE model. 
As the HOPE model becomes powerful, it can also continuously promote the current ArtiBoost to evolve.
ArtiBoost is model-agnostic, which means it can be plugged into any modern CNN architecture.
In this paper, we plug ArtiBoost into a simple classification-based (\eg \cite{sun2018integral}) and regression-based (\eg \cite{boukhayma20193d}) pose estimation model to show its efficacy. For evaluation, we report those models' performance on two challenging HOPE benchmarks: HO3D (v1-v3) \cite{hampali2020ho3dv1 ,hampali2020ho3dv2, hampali2021ho} and DexYCB \cite{chao2021dexycb}.
Without whistles and bells, those simple baseline models can outperform the results of previous state-of-the-arts. 

In this paper, we propose to boost the performance of HOPE task by enhancing the diversity of underlying poses distribution in the training data. We summarize our contributions as follows.
(1) To describe the composited hand-object-viewpoint poses distribution, we design the CCV-space. 
(2) To overcome the scarcity of such poses in the previous dataset, we design a contact-guided grasp synthesis method and simulate both valid and diverse hand-object poses to fill the CCV-space.
(3) To help HOPE model efficiently fit the underlying poses distribution, we parallelize the data synthesis with the learning pipeline, leverage the training feedback, and adopt a sample re-weighting strategy.
Finally, We conduct extensive experiments to validate our technical contributions (\cref{sec:hope_exp}, \ref{sec:ablation}) and reveal the potential applications (\cref{sec:application}).

\vspace{-1.0mm}\section{Related Work}
\vspace{-1.0mm}\noindent\textbf{Hand-Object Pose Estimation.}
As some HOPE tasks are closely related to hand pose estimation (HPE) tasks \cite{wan2019self, yang2021semihand}, 
we firstly review several HPE methods.
According to its output form, single RGB-based 3D HPE can be categorized into three types: image-to-pose (I2P) \cite{yang2019disentangling}, image-to-geometry (I2G) \cite{chen2021handmesh, tang2021handar}, and hybrid \cite{zhou2020monocular}.
While the I2P only focuses on the joints' pose only, the I2G focus on recovering the full hand geometry (pose and shape).
Meanwhile, recent works \cite{zhou2020monocular, yang2020bihand, li2021hybrik} showed that I2G could be hybridized to I2P through neural inverse kinematics (IK).
Second, we explore several HOPE methods. 
Regarding the learning-based methods, some aimed to predict the hand-object poses in a unified model \cite{hasson2020leveraging, liu2021semi}, while the others focused on recovering hand-object interaction based on contact modeling \cite{yang2021cpf, grady2021contactopt}. As for the learning-free methods, Hasson \etal \cite{hasson21homan} and Cao \etal \cite{rhoi2020} proposed to aggregate the visual cues from object detection, HPE, and instance segmentation to acquire the optimal hand-object configuration.  
This paper adopts two simple learning baseline networks of two paradigms: classification (joints as 3D heatmap) and regression (joints from pose and shape predictions). We show that with ArtiBoost, even simple baseline networks can outperform previous sophisticated CNN designs.

\vspace{1.0mm}\noindent\textbf{Data Synthesis for Pose Estimation.}
Using synthetic data to increase pose variants has been widely adopted in single-body N-D pose estimation tasks, such as human pose \cite{varol21_surreact, yan2021ultrapose, chen2016synthesizing}, hand pose \cite{zimmermann2017rhd, ge20193d, chen2021MobRecon}, 6D object pose \cite{kehl2017ssd6d, tekin2018SeamlessSingleShot, sundermeyer2018implicit}, and 7D articulated object pose \cite{li2020ancsh, liu2022towards}, \etc. 
These methods utilized the kinematic model of the articulated bodies, drive constrained joints' motion, and rendered the current model from different viewpoints.  
Unlike single-body pose estimation, Data synthesis for HOPE (multi-body) task not only needs to consider the joints limit, but also to obey the physical constraints brought from mutual contact and obstruction.
In terms of composited hand-object pose synthesis, there are also three genres included in the literature. 
The manual labeling methods \cite{romero2010_handsInAction, Rogez2015_GUN71, corona2020ganhand} articulated a hand model to achieve grasp;
The metric-based methods \cite{hasson2019obman, kokic2020learningTOG,brahmbhatt2019contactgrasp} leveraged grasp simulator \cite{graspit} to simulate hand poses subjected to physical grasping metrics; 
The data-driven methods \cite{taheri2020grab, jiang2021graspTTA, karunratanakul2020graspingfield} trained conditional VAE that generates new grasps. 
This paper presents an automatically contact-guided optimization method to construct valid and diverse poses of hand-object interaction.

\vspace{1.0mm}\noindent\textbf{Exploration for Hard Examples,} also called ``hard example$/$negative mining'', has been proven effective for various computer vision tasks, such as object detection \cite{hard_obj_det1, hard_obj_det2}, person re-id \cite{hard_person_reid1}, head pose estimation\cite{Kuhnke2019DeepHP}, face recognition \cite{hard_face1}, and deep metric learning \cite{hard_deep_metric1, hard_deep_metric3}. 
Generally speaking, the basic ideology of hard examples mining is that if a prediction of a certain data sample exhibits a large error under a certain metric, then this data sample is not properly learned by the learning algorithm. 
By adding such data samples to the training batch can help the learning converge faster. Recent work exploited a generative adversarial model \cite{gong2021poseaug} to acquire harder samples based on error feedback training. However, it must pay non-trivial efforts on the adversarial part to ensure samples' validity. 
In this paper, we adopt a simple yet effective sample re-weighting strategy that adaptively selects those hard-discernable training triplets (hand-object-viewpoint) for training.

\vspace{-2 mm}\section{Method}
\vspace{-1 mm}\noindent\textbf{Overview.}
This section describes the exploration and synthesis step in ArtiBoost and elaborates the learning framework for the HOPE task. 
Inside the explorations step, we present the composited configurations and viewpoints space (CCV-space), the key component of ArtiBoost. 
\begin{figure}[!t]
    \begin{center}
      \includegraphics[width=0.91\linewidth]{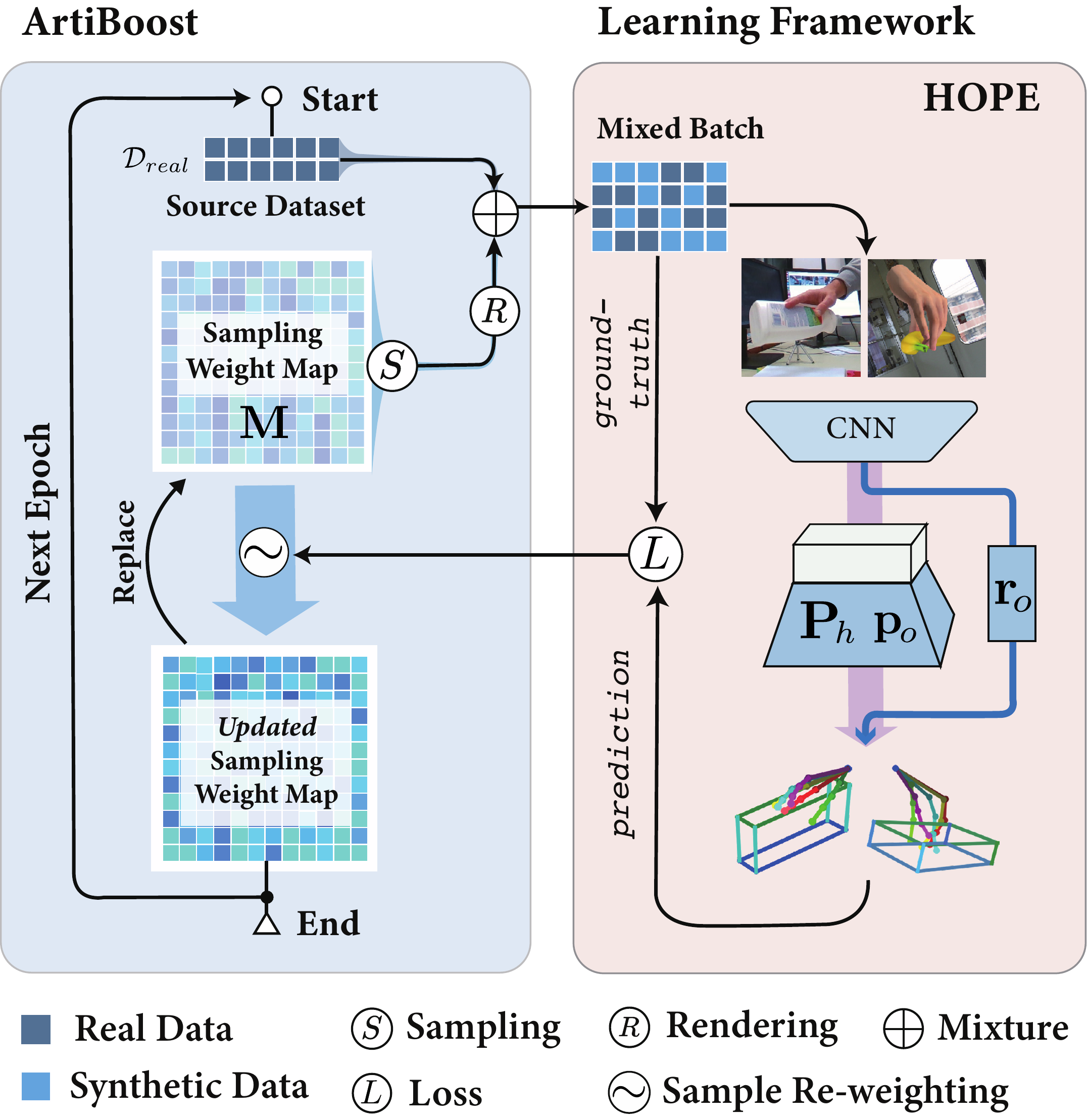}
    \end{center}\vspace{-6mm}
      \caption{\textbf{Illustration of the integrated pipeline.} ArtiBoost can be plugged into an arbitrary HOPE framework by modifying the current data loader.} \vspace{-3 mm}
    \label{fig:pipeline}
\end{figure}

\vspace{1 mm}\noindent\textbf{Problem Definition.}
Given an input image $\mathbf{I} \in \mathbb{R}^{H \times W \times 3}$ that observes a single hand interacting with a certain object, HOPE aims to learn a certain neural network that predict the 3D hand joint locations: $\mathbf{P}_h=\{\mathbf{p}_j\}^{J}_{j=1}$, object centroid locations: $\mathbf{p}_o$ and object rotation: $\mathbf{r}_o \in \mathfrak{so}(3)$,  where $\mathbf{p}_j, \mathbf{p}_o \in \mathbb{R}^3$, $J=21$ and the $H \times W$ is the resolution. 

To train the neural network, we shall firstly prepare a real-world source dataset: $\mathcal{D}_{real}$.
ArtiBoost is employed along with the $\mathcal{D}_{real}$.
During the training process, ArtiBoost iteratively samples (without replacement) hand-object-viewpoint triplets from the CCV-space: $\mathcal{C}$ based on a weight map: $\mathbf{M}$. 
Each entry of $\mathbf{M}$ corresponds to a certain triplet in $\mathcal{C}$, and the value in that entry corresponds to the sampling weight of the triplet.
Meanwhile, those selected triplets are rendered as a batch of synthetic images at the synthesis step.
The synthetic images are mixed with source images from $\mathcal{D}_{real}$.
After that, the mixed batch is fed to the HOPE learning framework to complete a forward and backward propagation. 
When an epoch of training has finished, 
ArtiBoost performs the sample re-weighting in $\mathbf{M}$ based on the loss value and waits for the next round of training. The whole pipeline is illustrated in \cref{fig:pipeline}.

\subsection{Online Exploration in CCV-Space}
\vspace{-1mm}\noindent\textbf{The Composited Configuration \& Viewpoint Space.}

\noindent HOPE problem commonly involves a certain interacting hand-object configuration that is observed by a certain viewpoint.
The input domain of HOPE can thus be narrowed down to three main dimensions: object type, hand pose, and viewpoint direction.
To note, the dimension of object type and hand pose are not independent of each other. 
Given a certain object model, the hand pose that interacts with it depends on the geometry of the model. 
As shown in \cref{fig:capture_ccvspace}, we define the discrete representation of the input domain as the CCV-space:
$\mathcal{S} = \{ (n_o, n_p, n_v) \in \mathbb{N}_+^3  \ | \ n_o \le N_o, \ n_p \le N_p, n_v \le N_v \}$,
where the $N_o$, $N_p$ and $N_v$ is the number of object types, discrete poses and viewpoints, respectively.
The $(i,j,k)$ item in $\mathcal{S}$ stands for the scenario that the interaction between the $i$-th object and the $j$-th hand pose is observed at the $k$-th camera viewpoint.
Next, we will sequentially present the components in CCV-space, namely: hand configuration space (\textbf{C-space}),  composited hand-object configuration space (\textbf{CC-space}) and  viewpoint space (\textbf{V-space}).

\begin{figure*}[ht]
    \begin{center}
        \begin{minipage}[ht]{0.18\textwidth}
        \centering
        \resizebox{0.93\linewidth}{!}{
            \includegraphics[width=1.0\linewidth]{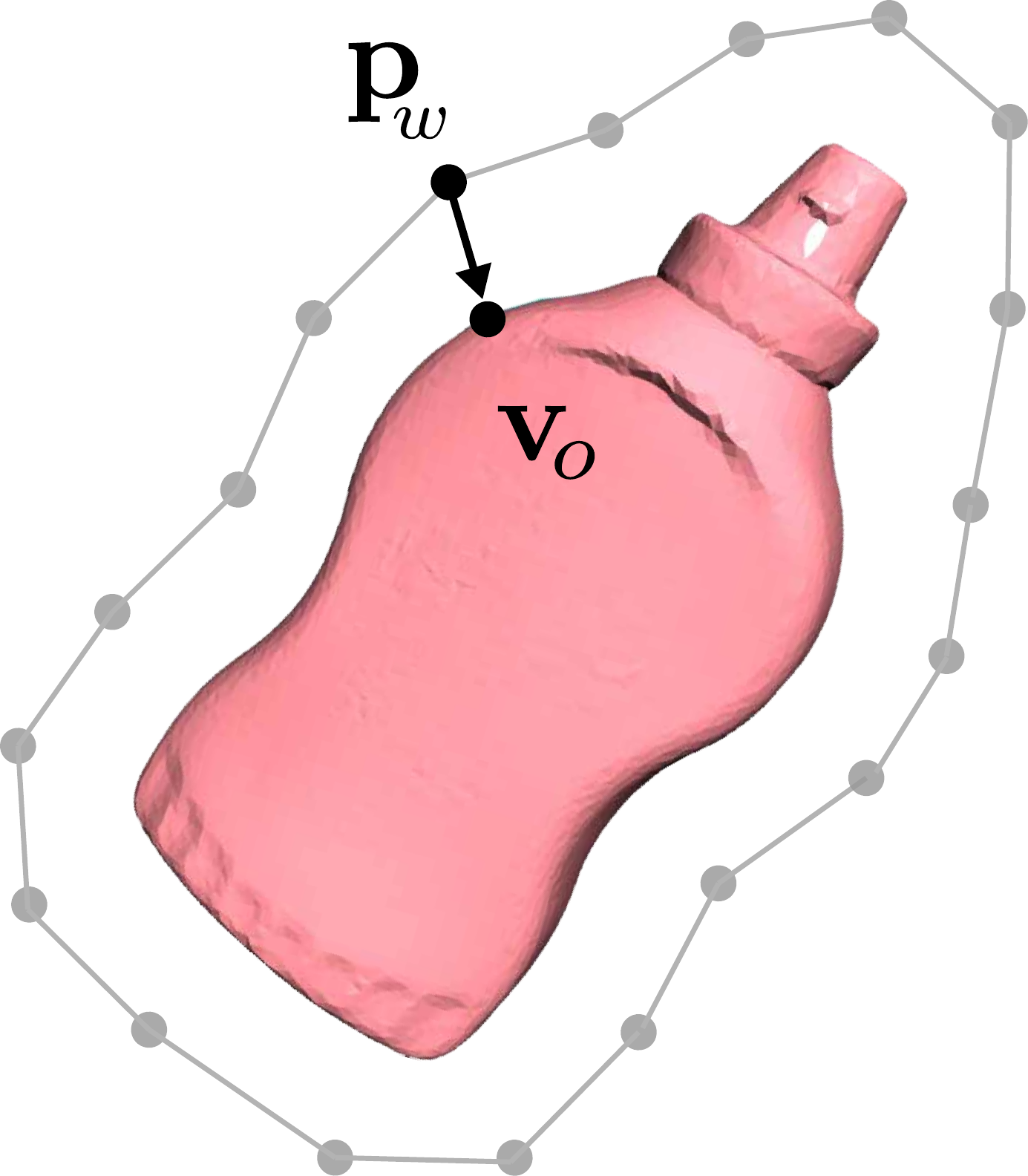}
            }
            \vspace{-0.1em}
            \caption{Offset surface.}
            \label{fig:offset_surface}
        \end{minipage}
        \quad
        \begin{minipage}[htp]{0.33\textwidth}
        \centering
            \resizebox{0.95\linewidth}{!}
            {
                \includegraphics[width=1.0\linewidth]{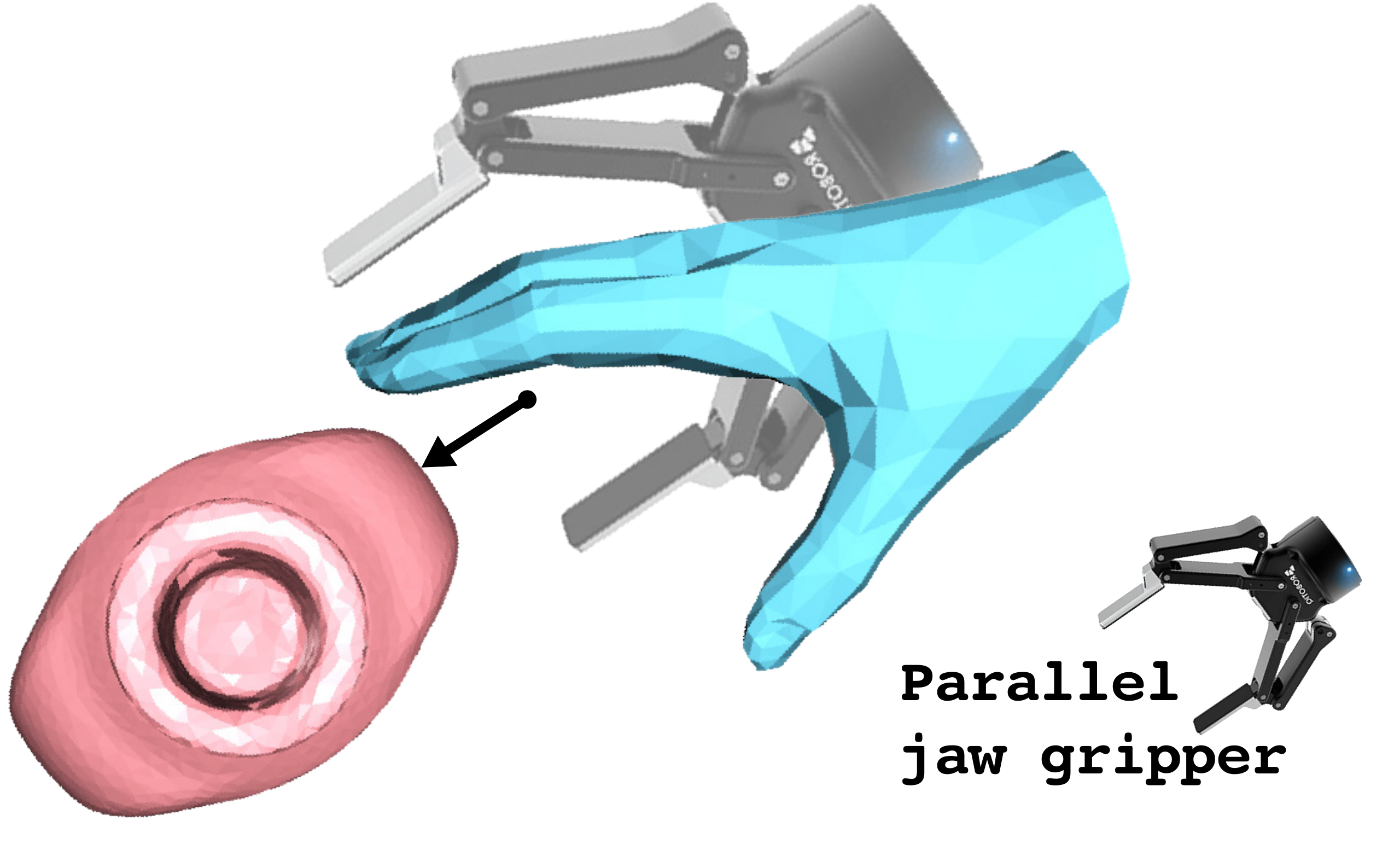}
            }
            \vspace{-1.0mm}
            \caption{Pre-grasping hand pose.}
            \label{fig:pregrasping_hand}
        \end{minipage}
        \quad
        \begin{minipage}[htp]{0.30\textwidth}
        \centering
            \resizebox{0.87\linewidth}{!}
            {
                \includegraphics[width=1.0\linewidth]{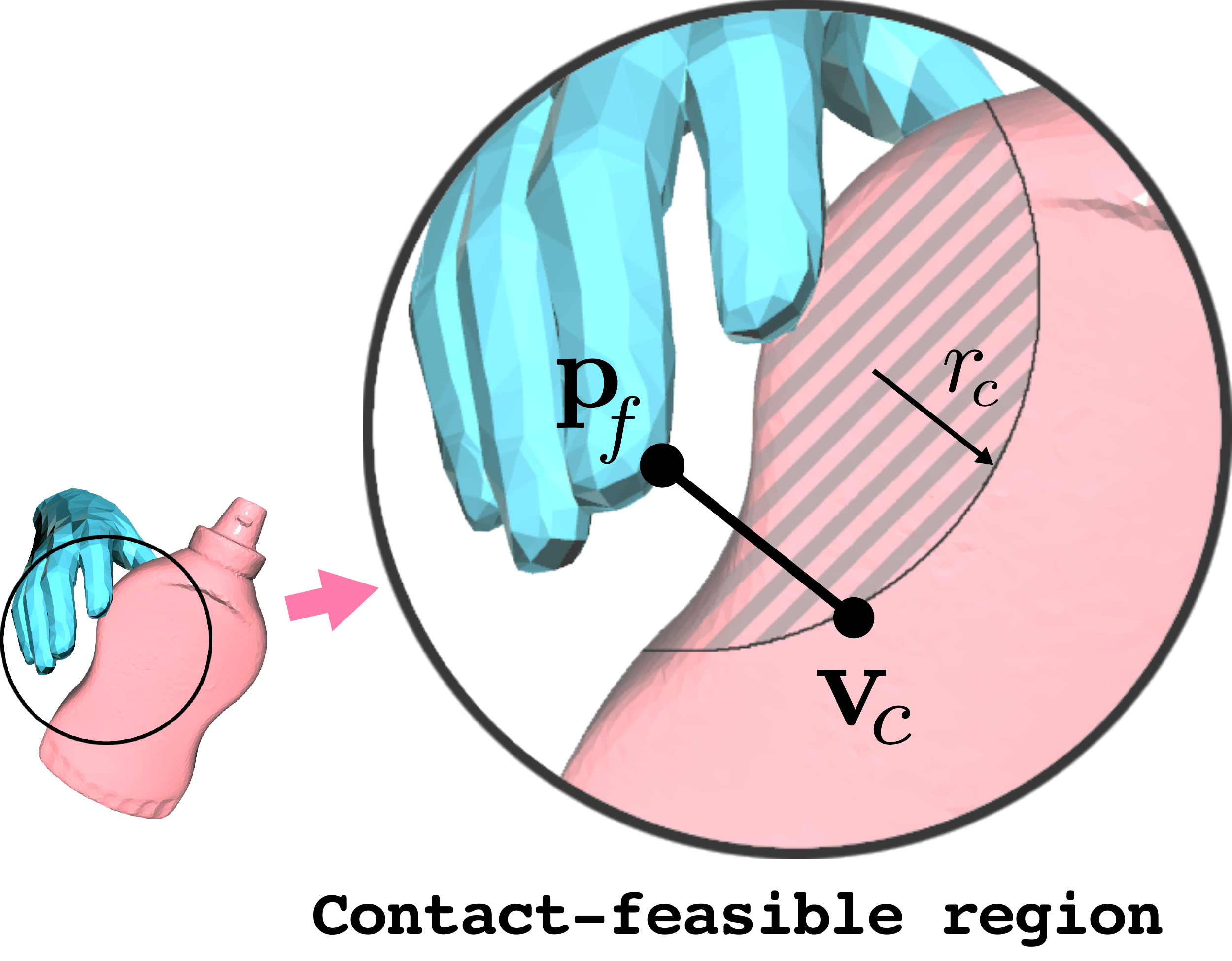}
            }\vspace{-2.5mm}
            \caption{Fingertip and its contact point.}
            \label{fig:paring_fingertips}
        \end{minipage}\vspace{-7.0mm}
    \end{center}
\end{figure*}

% ##################################################
% #######        Hand C-space      #################
% ##################################################
\vspace{1mm}\noindent\textbf{C-Space of Valid Hand Pose.}
To represent hand, we employ a parametric skinning hand model, MANO \cite{romero2017embodied} which drives a deformable hand mesh with 16 joints rotations $\bm{\theta} \in \mathbb{R}^{16 \times 3}$ and shape parameters $\bm{\beta} \in \mathbb{R}^{10}$. Given the axis-angle forms of rotation, MANO has 48 DoFs that exceed the DoFs allowed by a valid hand pose \cite{lin2000constraintshand}. 
Fitting or interpolation on the 48 DoFs rotations may encounter abnormal hand pose that is unhealthy for training the HOPE network.
Besides, the original MANO's coordinate system is not coaxial with the direction of the hand's kinematic tree, so that the rotation axis is coupled with at least two of the MANO's orthogonal axes. 
This property makes the pose interpolation more difficult.
In the paper, we employ an axis adaptation based on the \textit{twist-splay-bend} 
coordinate system that was initially proposed by \cite{yang2021cpf}. 
It enables us to describe the hand pose at each joint as the rotation angle along with one of the specified coordinate axis (\eg \textit{bend}) at this joint.
With the axis adaptation, we design three protocols for describing the C-space of valid hand pose:
\begin{enumerate}[label={\roman*).},font={\bfseries}]
    \setlength\itemsep{-0.5 mm}
    \vspace{-1.0mm}
    \item According to hand anatomy, all the non-metacarpal joints along the hand's kinematic tree can only have the bending pose. And the five metacarpal joints can only have a combined bending and splaying poses.
    Any twisting along the pointing direction or splaying at non-metacarpal joints is prohibited.
    \item For each of the five fingers, the bending poses on their proximal and distal joint are linked and dependent. 
    The bending poses at the five metacarpal joints are independent of other joints.
    \item The pose of each finger is independent of each other as long as it does not conflict with the protocol \textbf{i)} \& \textbf{ii)}.\vspace{-1.0mm}
\end{enumerate}
Based on these protocols, the total DoFs in the hand's C-space is 21 (one splaying and two independent bending DoFs for each of the five fingers, plus 6 DoFs at the wrist).
Hence, we can describe the whole hand pose by describing the angles of 15 joints' rotation along its specified axis: $\theta_{i}^{\textit{ bend}}$ or $\theta_i^{\textit{ splay}}$, as well as the wrist pose: $\bm{\xi}_{w} \in \mathfrak{se}(3)$.
These protocols not only guarantee the diverse and valid hand poses but also ensure these poses are visually plausible.
After constructing a valid C-space, we now move on to the composited space that describes hand-object interaction. 

% ##################################################
% #######   Hand-object CC-space   #################
% ##################################################

\vspace{1mm}\noindent\textbf{Composited C-Space of Hand-Object Interaction.}
We firstly define the hand-object ``interaction'' as the scenario that satisfies the following requirements: 
\begin{enumerate*}[label={\roman*)},font={\bfseries}]
    \item The thumb and at least one or more fingers should be in contact with the object surface \cite{taheri2020grab};
    \item The hand and object model should not intersecting with each other;
\end{enumerate*}
As the hand's interacting pose is highly dependent on its approaching direction and the object's geometry, only a small portion of poses in hand C-space is valid for interaction purposes. 
Hence instead of searching for valid interaction poses in the entire hand C-space, we turn to explore a discrete space of predefined hand-object poses. We call it composited C-space (CC-space).
The goal of CC-space is to increase the diversity of the interacting poses for a given object. 
For this purpose, 
we leverage the contact constraints to optimize grasps. This method is divided into three steps.

\vspace{0.5mm}\textbf{$\bullet$~1).~}First, given an object model, we construct an offset surface outside of its original surface and uniformly sample $N_w$ points on the offset surface (\cref{fig:offset_surface}). These points will control the wrist position. For each point: $\mathbf{p}_w$, we query its closet vertex on object surface: $\mathbf{v}_o$. The vector: $(\mathbf{v}_o-\mathbf{p}_w)$ represents the approaching direction from hand to object. Then, we construct a prehensile hand pose that mimics the rest state of a parallel jaw gripper (\cref{fig:pregrasping_hand}).
    This pre-grasping hand is placed at each $\mathbf{p}_w$ as interaction initiation. The $\mathbf{p}_w$'s approaching direction controls the movement of the wrist. 
    Based on a certain $\mathbf{p}_w$, we define the region between $\mathbf{v}_o$ and the farthest vertex that the finger can reach as the object's contact-feasible region. 

\vspace{0.5mm}\textbf{$\bullet$~2).~}Second, to generate the interacting pose, we fit the fingertips to contact point $\mathbf{v}_c$ chosen from contact-feasible region (\cref{fig:paring_fingertips}). 
For each $\mathbf{p}_w$, we generate $N^{w}_{p}$ interacting hand poses.
To increase diversity, we employ several randomness during fitting: 
\begin{enumerate*}[label={\roman*)},font={\bfseries}]
    \item We randomly select thumb and $N$ $(1 \le N \le 4)$ other fingers on hand. Only the selected fingers will participate in fitting.
    \item For each selected fingers, we set a random minimal reaching radius $r_c$ inside the contact-feasible region. 
\end{enumerate*}
Now the selected fingertip must reach for a contact point $\mathbf{v}_c$ that satisfies:
\begin{enumerate*}[label={\roman*)},font={\bfseries}]
    \item $\|\mathbf{v}_c - \mathbf{p}_w \|_2 \ge r_c$; 
    \item ${\min_{\mathbf{v}_c}}\|\mathbf{v}_c - \mathbf{p}_f \|_2$, where the $\mathbf{p}_f$ is the selected fingertip point.  
\end{enumerate*}
After each selected fingertip is paired with a certain contact point, we will initiate the fitting process. 
    
\vspace{0.5mm}\textbf{$\bullet$~3).~}During fitting, we adopt the anchor-based hand model and contact-based cost function defined in CPF \cite{yang2021cpf}. The fitting process aims to minimize the cost brought from unattached fingertips and contact points: the unattached anchor on a fingertip: $\mathbf{p}_f$ will be attracted to its corresponded contact point $\mathbf{v}_c$ on the object, while the intersected anchors will be pushed out. The fitting of hand poses is only performed on the predefined DoFs in the hand's C-space. Thus valid hand pose can be guaranteed. 

% ##################################################
% #######      Viewpoint space     #################
% ##################################################
\vspace{1 mm}\noindent\textbf{Viewpoint Space.}
For the camera viewpoint, we uniformly sample ${N_{v}}$ viewpoints direction $\mathbf{n}_{v}$ according to the sphere sampling strategy \cite{marsaglia1972choosing}:
\begin{equation}
    \setlength\abovedisplayskip{5pt} %shrink space
    \setlength\belowdisplayskip{5pt}
    \mathbf{n}_v = (\sqrt{1-u^2} \cos(\phi), \sqrt{1-u^2} \sin(\phi), u)^\mathsf{T}
\label{eq:sphere_sampling}
\end{equation}
where $u \sim \mathcal{U}[-1, 1]$ and $\phi \sim \mathcal{U}[0, 2\pi]$, where $\mathcal{U}$ stands for uniform distribution. 

\vspace{0.5 mm}\noindent\textbf{CCV-space Implementation.}
We iteratively fit 300 different interacting poses for each object. Since optimization may result in local minima, we manually discard the poses that 
\begin{enumerate*}[label={\roman*)},font={\bfseries}]
    \item exhibit severe inter-penetration between hand and object;
    \item form an unnatural grasping or interaction.
\end{enumerate*}
Then, we select up to 100 interacting hand poses per given object. 
Though the optimization can potentially generate unlimited interacting poses, we find 100 poses (per object) are 
sufficient to boost current HOPE tasks.
In the viewpoints space, we choose $N_u =12$ and $N_{\phi} = 24$, which comprise $N_v = N_u \times N_{\phi} = 288$ different viewpoints. 
The total number of different (hand-object-viewpoint) triplets depends on the benchmarking dataset. For example, the triplets catered for DexYCB dataset (containing 20 YCB objects) is $N_o \times N_p \times N_v = 20 \times 100 \times 288 = 576,000$.

\vspace{1 mm}\noindent\textbf{Weight-guided Sampling Strategy.}
In literature, the uniform sampling strategy was widely adopted by synthetic dataset \cite{zimmermann2017rhd, hasson2019obman, mueller2018ganerated}. However, not every sample in the CCV-space contributes equally to the network. Since we hope that those hardly discernable samples shall have a higher frequency of occurrence, we construct a weight map $\mathbf{M} \in \mathbb{R}^{N_o \times N_p \times N_v}$ to guide the sampling in the exploration step. In $\mathbf{M}$, each element $w_i$ stands for the sampling weight of the corresponding item in CCV-space. The probability $p_i$ of a certain item that would be sampled is $p_i = w_i / \sum{w_j}$.
We then draws $N_{syn}$ samples from the multinomial distribution $\{ p_i \ | \ p_i = w_i / \sum{w_j};\ \ w_i, w_j \in \mathbf{M} \}$.
Based on this strategy,
we shall increase the weights for those hardly discernable samples in $\mathbf{M}$ while decreasing the weights for those who are already easy to discern, when we get the feedback from the loss value.

\vspace{1 mm}\noindent\textbf{Sample Re-weighting.}
After an epoch of training has finished, ArtiBoost chooses those hard discernable items and re-weights their sampling weights.
We inspect a percentile-based re-weighting strategy.
During the re-weighting phase, each synthetic sample will be assigned a weight update.
These updates are multiplied by the original sampling weight in $\mathbf{M}$.
Intuitively, we want those hard discernable samples to have high weight. 
In the percentile-based re-weighting strategy, we calculate the weight update based on the percentile of the samples' Mean Per Joint Position Error (MPJPE) among the whole epoch of synthetic samples.
For the $i$-th sample, given by the MPJPE $e_i$ and its percentile $q_i = \frac{e_{\text{max}} - e_i}{e_{\text{max}} - e_{\text{min}}}$, the weight update are calculated from a simple reciprocal heuristic: $\delta w_i = \frac{1}{q_i + 0.5}$. If the sample $i$ has the maximum MPJPE $e_{\text{max}}$ among the synthetic samples in current epoch, its original sampling weight in $\mathbf{M}$ will be multiplied by a maximum update factor $\delta w_i = 2$. If the $i$ has the minimum MPJPE $e_{\text{min}}$, its update factor would be $\delta w_i = 2/3$. We also clamp the updated $\mathbf{M}$ by a upper bound 2.0 and lower bound 0.1 to avoid over imbalance.

\vspace{-1 mm}\subsection{Online Synthesis for HOPE task}\label{sec:rendering}
\vspace{-1 mm}During the training process, we synthesize the sampled hand-object-viewpoint triplets to RGB images. 
This synthesis process is task-oriented, as the adaptive sampling decides its composition to cater to the downstream task. 
Here, we describe the features in the online synthesis step.

\vspace{1 mm}\noindent\textbf{Disturbance on the Triplets.} 
To increase the variance in the pose distribution and thus to improve the network's generalization ability, we add disturbance on the hand poses and viewpoint directions before rendering images.

\textit{$\bullet$~~For the hand poses}, 
we relieve the restriction in protocol \textbf{ii)} in hand C-space, in which the bending angles of distal and proximal joints on each finger are now independent in terms of disturbance. 
Then, we add a Gaussian disturbance $\mathcal{N}(0, \sigma_1^2)$ on each of the 15 bending angles.
Second, for the disturbance on splaying angles, we add a $\mathcal{N}(0, \sigma_2^2)$ on the five metacarpal joints. We empirically set $\sigma_1 = 3$ and $\sigma_2 = 1.5$ degree. 
To note, this disturbance still subject to the restrictions in protocol \textbf{i)} and \textbf{iii)}, which ensure a valid and prehensile hand configuration.
However, these disturbances may cause the inter-penetration between the hand and object models. 
Hence, we further process the disturbed hand-object pose through the \textit{RefineNet} module in GrabNet \cite{taheri2020grab}, which has the effect of mitigating inter-penetration.  
Apart from hand pose, the shape of hand also impacts the interaction with the object.
Hence we sample the random MANO shape parameters ($\beta \in \mathbb{R}^{10}$) from the distribution of $\mathcal{N}(0, 0.5)$ to formulate the final hand model.

\textit{$\bullet$~For the viewpoint directions}, we add three disturbances: $\mathcal{U}(-\delta u, + \delta u)$, $ \mathcal{U}(-\delta \phi, +\delta \phi)$, and $ \mathcal{U}(0, 2\pi)$ on the elevation distance $u$, azimuth angle $\phi$ and the camera in-plane rotation, respectively.
In all experiments, we empirically set $\delta u = 0.05$ and $\delta \phi = 7.5$ degree.

\vspace{1 mm}\noindent\textbf{Skin Tone \& Textures.} We adopt a state-of-the-art hand skin tone \& texture model: HTML \cite{qian2020html} for realistic appearance on the rendered images.
HTML represents the hand's skin color \& texture of as continuous parameters in a PCA space.
Before the refined hand model enters the rendering pipeline, we randomly assign it an HTML texture map.
We also found that discarding the shadow removal operation in HTML produces more visually plausible images. 

\begin{figure}[t]
    \includegraphics[width=1.0\linewidth]{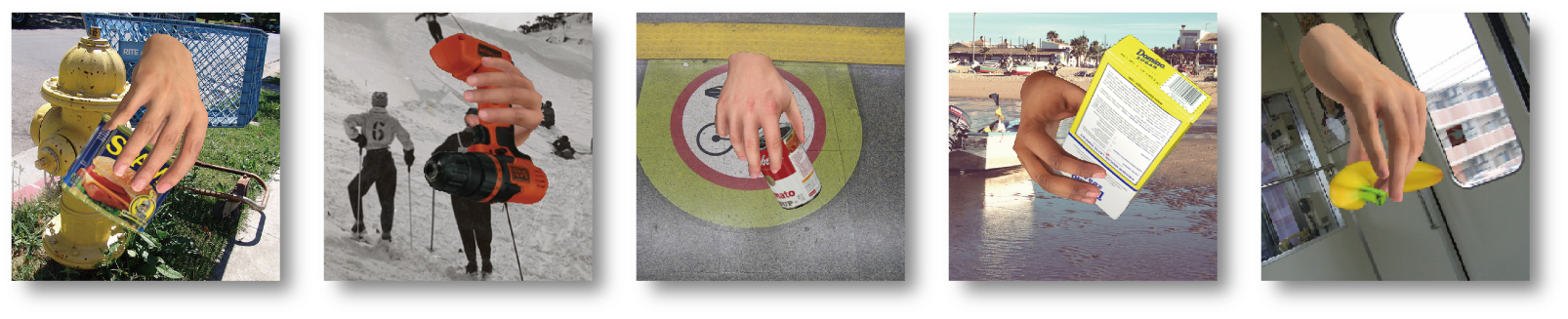}\vspace{-3mm}
       \caption{The rendered images during online synthesis.}\vspace{-6mm}
    \label{fig:dataset_vis}
\end{figure}

\vspace{1 mm}\noindent\textbf{Rendering.} We employ the off-the-shelf rendering software: PyRender \cite{pyrender} inside ArtiBoost.
The interactive hand and object are the foreground, and the images in COCO \cite{lin2014microsoft} dataset are the background. 
The pipeline can support rendering of the photorealistic hand and object textured meshes onto a $224 \times 224$ canvas at 120 FPS per graphic card (Titan X), sufficient for us to parallelize the rendering with training. 
The synthetic data is the combination of four customized items, namely hand-object-viewpoint triplet, background, skin tone, and texture. We show several synthetic images in \cref{fig:dataset_vis}.

\vspace{-1mm}\subsection{Learning Framework}\label{sec:framework}
\vspace{-2 mm}We adopt two simple baseline network: one is classification-based (\textit{Clas}) and the other regression-based (\textit{Reg}). 
We employ ResNet-34 \cite{he2016deep} as the backbone in both of them.
In \textit{Clas}, we use two de-convolution layers to generates 22 3D-heatmaps 
that indicate the location of 22 joints (21 hand joints and one object centroid) as likelihood.
The 22 3D-heatmaps are defined in a restricted \textit{uvd} space, where \textit{uv} is the pixel coordinates, and \textit{d} is the a wrist-relative depth value.
Then, we employ a soft-argmax operator to convert the 3D-heatmaps into joints' \textit{uvd} coordinates.
Finally, we transform the joints' \textit{uvd} coordinates into its 3D locations: $\mathbf{P}_h, \mathbf{p}_o$ in camera space by camera intrinsic: $\mathbf{K}$ and wrist location: $\mathbf{p}_w$.
In \textit{Reg}, we use multi-layer perceptron (MLP) to regress the MANO parameters: $\bm{\theta}$, $\bm{\beta}$ and object centroid: $^w\mathbf{p}_o$ w.r.t. the wrist. Then we transfer the $\bm{\theta}$, $\bm{\beta}$ to the wrist-relative hand joints: $^w\mathbf{P}_h$ by the MANO model. Finally the  $^w\mathbf{P}_h$ and  $^w\mathbf{p}_o$ are translated into the camera space by adding a known wrist location: $\mathbf{p}_w$.  Both \textit{Clas} and \textit{Reg} adopt another MLP branch to predict object rotation: $\mathbf{r}_o$. 
Detailed implementations are provided in \suppl.

\vspace{1mm}\noindent\textbf{Loss Function.}
The loss function to train the HOPE network consists of four terms. 

First, we penalize the error of total 22 joints location (21 hand joints and 1 object centroid) in form of $\ell 2$ distance: 
\begin{equation}
    \setlength\abovedisplayskip{5pt} %shrink space
    \setlength\belowdisplayskip{5pt}
    \mathcal{L}_{loc} = \frac{1}{22} \sum_{i=1}^{22} \Big \| \mathbf{p}_i - \mathbf{\hat{p}}_i \Big \|^2_2
\label{eq:loc_loss}
\end{equation} where $\mathbf{\hat{p}}_i$ denotes the ground-truth joint location.

Second, we penalize the error of object rotation in form of $\ell 2$ distance at the eight tightest bounding box corners:
\begin{equation}
    \setlength\abovedisplayskip{5pt} %shrink space
    \setlength\belowdisplayskip{5pt}
    \mathcal{L}_{cor} = \frac{1}{8} \sum_{i=1}^{8} \Big \| \exp(\mathbf{r}_o) * \mathbf{\mathbf{\bar{c}}}_i  - \mathbf{\hat{c}}_i \Big \|^2_2
\label{eq:cor_loss}
\end{equation}
where the $\mathbf{\bar{c}}_i$ and $\mathbf{\hat{c}}_i$ are the object's corners in canonical view and camera view. $\exp(\mathbf{r}_o)$ is the predicted rotation.

Third, we adopt the ordinal relation loss to correct the 2D-3D misalignment. 
$\mathcal{L}_{ord}$ inspects the joint-level depth relation inside a pair of joints: 
one from the 21 hand joints and the other from the 8 object corners. 
We penalize the case if the predicted depth relation between 
the $i$-th hand joint: $\mathbf{{p}_i}$ and the $j$-th corner $\mathbf{{c}_j}$ is misaligned with its ground-truth relation: $\mathds{1}^{ord}_{i,j}$.
The $\mathcal{L}_{ord}$ is formulated as:
\begin{equation}
    \setlength\abovedisplayskip{5pt} %shrink space
    \setlength\belowdisplayskip{5pt}
    \mathcal{L}_{ord} =  \sum_{j=1}^{8}\sum_{i=1}^{J=21} \mathds{1}^{ord}_{i,j} * \Big | (\mathbf{p}_i - \mathbf{c}_j) \cdot \mathbf{n}_{\perp} \Big |
\label{eq:ord_loss}
\end{equation}
where the $\mathbf{n}_{\perp}$ is the viewpoint direction.

Fourth, we borrow a symmetry-aware object corner loss: $\mathcal{L}_{sym}$ from Hampali \etal \cite{hampali2021handsformer}:
\begin{equation}
    \setlength\abovedisplayskip{5pt} %shrink space
    \setlength\belowdisplayskip{5pt}
    \mathcal{L}_{sym} = \min_{\mathbf{R} \in \mathcal{S}} \frac{1}{8} \sum_{i=1}^{8} \Big \| \exp(\mathbf{r}_o)* \mathbf{\mathbf{\bar{c}}}_i  -  \exp(\mathbf{\hat{r}}_o) \mathbf{R} * \mathbf{\mathbf{\bar{c}}}_i \Big \|^2_2
\label{eq:sym_loss}
\end{equation}
where $\exp(\mathbf{\hat{r}}_o)$ denotes object's ground-truth rotation matrix. 
Given an object, the set $\mathcal{S}$ contains all the valid rotation matrices based on the object's predefined symmetry axes.

The overall loss is a weighted sum of the four terms:
\begin{equation}
    \setlength\abovedisplayskip{5pt} %shrink space
    \setlength\belowdisplayskip{5pt}
    \mathcal{L}_{\textit{HOPE}} = \mathcal{L}_{loc} + \lambda_1 \mathcal{L}_{cor} + \lambda_2\mathcal{L}_{ord} + \lambda_3\mathcal{L}_{sym}
\label{eq:overall_loss}
\end{equation}
where the $\lambda_{1\sim3}$ are the hyper-parameters.

\vspace{-1.6 mm}\section{Experiment and Result}\label{sec:exp}
\begin{table*}[ht]
    \begin{center}
        % Complexity
        \begin{minipage}[!htp]{0.3\textwidth}
        \begin{center}
        \makeatletter\def\@captype{table}\makeatother
        \resizebox{\linewidth}{!}{
            \begin{tabular}{l|c|c}
            \toprule
                Method & MPJPE & MPCPE \\
            \midrule
                Hasson \etal \cite{hasson2020leveraging} & 11.33 & 28.42 \\
                % \midrule
                Our \textit{Clas} w/o $\mathcal{L}_{ord}$ & 8.71 & \textbf{18.64}\\
                Our \textit{Clas} & \textbf{8.60} & 19.45\\
            \bottomrule
            \end{tabular}
            }
        \end{center}\vspace{-4.0mm}
        \caption{Comparisons with SOTA on \textbf{FHAB} dataset (errors are reported in $mm$). The comparisons are made in the wrist-aligned coordinates system. }
        \label{table:fhb_res}
        \end{minipage}
        ~~~
        \begin{minipage}[!htp]{0.23\textwidth}
        \begin{center}
            \resizebox{1.0\linewidth}{!}{
            \makeatletter\def\@captype{table}\makeatother
            \setlength{\tabcolsep}{1mm}{
                \begin{tabular}{l|c|c}
                \toprule
                Method &  MPJPE & MPCPE \\
                \midrule
                Hasson \etal \cite{hasson2020leveraging} & 3.69 & 12.38 \\
                Liu \etal \cite{liu2021semi} & 2.93 & - \\
                \midrule
                Our \textit{Reg} & 3.53 & 7.38\\
                Our \textit{Reg} \textbf{+ Arti} & \textbf{3.17} & \textbf{5.87}\\
                \midrule
                Our \textit{Clas} & 3.06 & 7.24\\
                Our \textit{Clas} \textbf{+ Arti} & \textbf{2.64} & \textbf{5.16}\\
                \bottomrule
                \end{tabular}
            }
        }
        \end{center}\vspace{-5.mm}
            \caption{Comparisons ($cm$) with SOTA on \textbf{HO3D} dataset.}
            \label{table:ho3d_res}
        \end{minipage}
        ~~~
        \begin{minipage}[!htp]{0.41\textwidth}
        \begin{center}
            \resizebox{1.0\linewidth}{!}
            {
            \makeatletter\def\@captype{table}\makeatother
            \setlength{\tabcolsep}{1mm}{
                \begin{tabular}{l|c|ccc}
                \toprule
                    \multirow{3}{*}{Method} & \multirow{3}{*}{MPJPE} & \multicolumn{3}{c}{MSSD} \\
                    \cline{3-5}
                    & & \tabincell{c}{\textit{mustard} \\ \textit{bottle}} & \tabincell{c}{\textit{bleach} \\ \textit{cleanser} } & \tabincell{c}{\textit{potted} \\\textit{meat can} } \\
                \midrule
                    Hampali \etal \cite{hampali2021handsformer} & 2.57 & 4.41 & 6.03 & 9.08 \\
                    Our \textit{Clas} sym & 3.10 & 4.07 & 6.56 & 8.70 \\
                    Our \textit{Clas} sym \textbf{+ Arti} & \textbf{2.53} & \textbf{3.14} & \textbf{5.72} & \textbf{6.36} \\
                \bottomrule
                \end{tabular}
                }
            }
        \end{center}\vspace{-6.0mm}
            \caption{Comparison ($cm$) with Transformer-based SOTA on \textbf{HO3D} using symmetry-aware loss $\mathcal{L}_{sym}$. We use the same symmetry axes as described in \cite{hampali2021handsformer}.} 
            \label{table:sym_ho3d_res}
        \end{minipage}

    \end{center}\vspace{-8mm}
\end{table*}

\vspace{-1mm}\subsection{Dataset and Metrics}
\vspace{-2mm}\noindent\textbf{Dataset. }
We evaluate our methods on three hand-object dataset: \textbf{FHAB} \cite{garcia2018first}, \textbf{HO3D} \cite{hampali2020ho3dv2} and \textbf{DexYCB} \cite{chao2021dexycb}. 
FHAB contains 20K samples of hand in manipulation with objects. We follow the ``action'' split as in Tekin \etal \cite{tekin2019h+o}, which contains 10,503 training and 10,998 testing samples. The FHAB dataset only contains a few numbers of hand poses and viewpoints. We find its training set is adequate for the neural network. Thus we only use FHAB to verify the feasibility of the learning framework. HO3D is a dataset that contains a large number of images of hand-object interactions. Evaluation of the HO3D testing set is conducted at an online server.  We also report our results on the latest \textbf{HO3Dv3} \cite{hampali2021ho}, which is released with different training$/$testing split. DexYCB contains 582K image frames of grasping on 20 YCB objects. We only evaluate the right-hand pose using the official ``S0'' split and filter out the frames that the minimum hand-object distance is large than 5 $cm$ 
to make sure a plausible hand-object interaction would appear.

\vspace{1mm}\noindent\textbf{Metrics.~~}
For the hand pose, we report the mean per joint position error (\textbf{MPJPE}) in the wrist-aligned coordinates system.
For the object pose, there are two standard metrics in literature: mean per corners position error (\textbf{MPCPE}) and maximum symmetry-aware surface distance (\textbf{MSSD}). The former MPCPE directly measures the unique pose of the object. 
However, since some objects are symmetrical or revolutionary invariant, and since the objects are often severely occluded by hand, direct measuring objects' absolute and unique pose is sometimes less reasonable. 
MSSD measures the difference between the current object pose to its closest counterpart in all its rotation invariants.
In this paper, we report the object's MPCPE and MSSD within different training schemes.
When reporting MPCPE, we train the network with $\lambda_1 = \lambda_2 = 1$ and $\lambda_3 = 0$. 
When taking objects' symmetricity into account, we train the network with $\lambda_1 = \lambda_2 = 0$ and $\lambda_3 = 1$.
We call the later network symmetry model (abbr. sym) and report its MSSD following the BOP challenge protocol \cite{hodavn2020bop}.  
The definition of YCB objects' symmetry axes can be found in \suppl.

\begin{figure}[t]
    \begin{center}
      \includegraphics[width=1.0\linewidth]{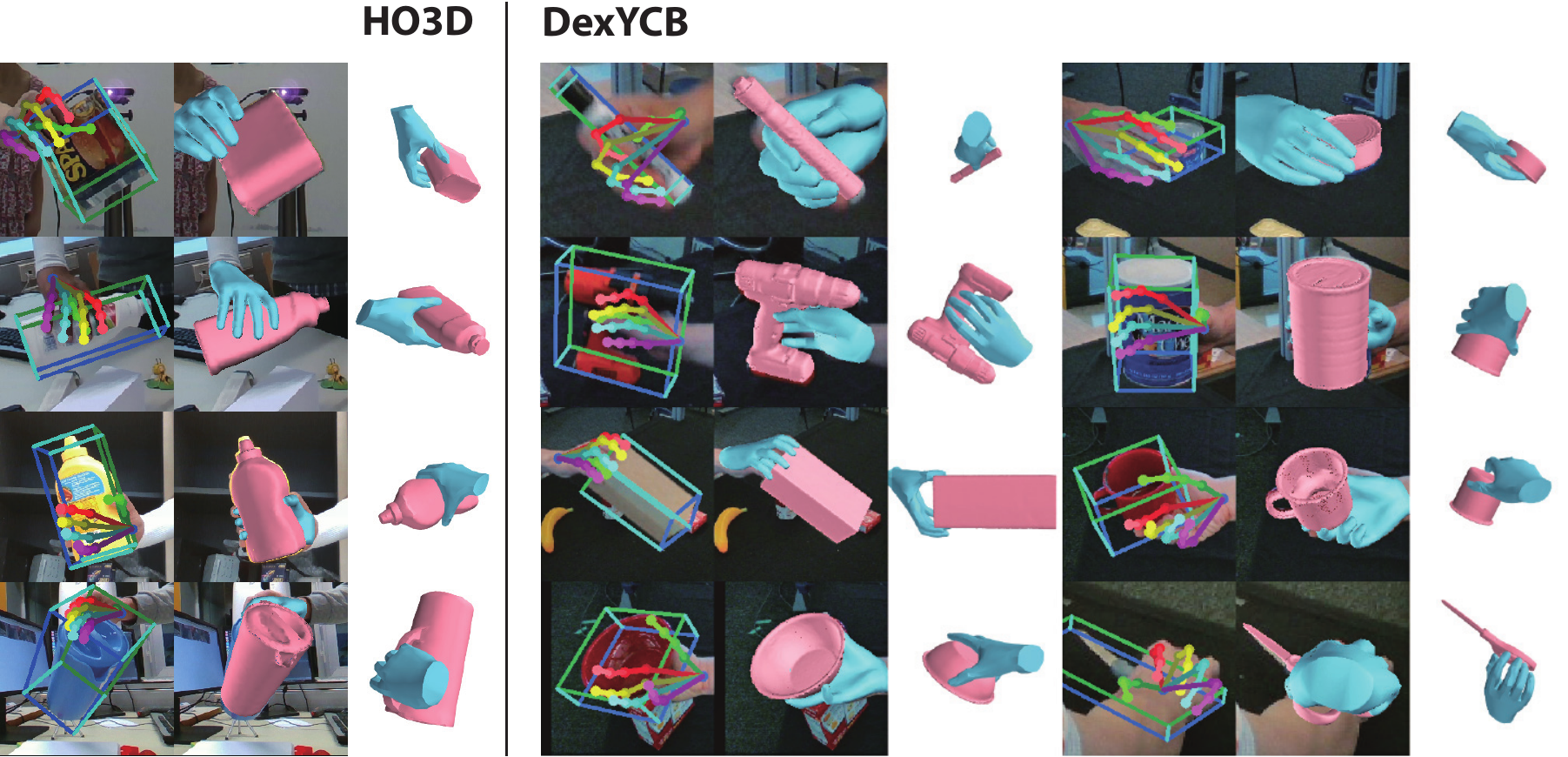}
    \end{center}\vspace{-5mm}
      \caption{Qualitative results on the  \textbf{HO3D} and \textbf{DexYCB} dataset.} \vspace{-4mm}
    \label{fig:qua}
\end{figure}
\subsection{HOPE Network Performance}\label{sec:hope_exp}
\vspace{-1mm}\noindent\textbf{Qualitative Results.} Our qualitative results on HO3D and DexYCB testing sets are shown in \cref{fig:qua}. We draw the predicted hand joints and object corners as their 2D projections (\engordnumber{1} col).
We also adopt a pretrained IK network \cite{lv2021handtailor} that maps the hand joints to hand mesh surfaces for visualization.
We draw the full hand-object geometry in camera view (\engordnumber{2} col) and another viewpoint (\engordnumber{3} col). More qualitative results are provided in \suppl. 

\vspace{1 mm}\noindent\textbf{Comparison with State-of-the-Art.} In \cref{table:fhb_res} we compare our \textit{Clas} with the previous SOTA \cite{hasson2020leveraging} on FHAB to justify the ordinal relation loss $\mathcal{L}_{ord}$. To note, \cite{hasson2020leveraging} regressed hand and object poses in camera space. For fair comparison, we align their results in wrist-relative coordinates system. 

In the subsequent tables, we use ``\textbf{+ Arti}'' to denote a certain network is trained by the original dataset's training split \textbf{plus} the synthetic data brought from \textbf{Arti}Boost's exploration and synthesis.
In \cref{table:ho3d_res}, we show that our ArtiBoost enhances both \textit{Reg} and \textit{Clas} performance on HO3D dataset.
We obtain \textbf{28\%} and \textbf{10\%} MPJPE improvement compared with the previous SOTA \cite{hasson2020leveraging} and \cite{liu2021semi}, respectively. 
For fair compassion, we remove the unseen object in the testing set (\textit{pitcher base}) when calculating MPCPE.

Under the symmetry model (denoted as ``sym''), we report the performance of our ArtiBoost in \cref{table:sym_ho3d_res}. Our \textit{Clas} outperforms the recent Transformer-based method \cite{hampali2021handsformer} when using ArtiBoost. We also evaluate our method on latest released dataset DexYCB and HO3D v3 in \cref{table:dexycb_res} and \cref{table:ho3dv3_res}, respectively. All the results demonstrate the effectiveness of our method.

\begin{table}[t]
    \renewcommand{\arraystretch}{1.0}
    \begin{center}
    \resizebox{\linewidth}{!}{
        \begin{tabular}{l|c|cccc}
        \toprule
            \multirow{3}{*}{Method} & \multirow{3}{*}{MPJPE} & \multicolumn{4}{c}{MSSD $\bm{\star}$} \\ \cline{3-6}
            & & \tabincell{c}{\textit{power} \\ \textit{drill}} & \tabincell{c}{\textit{cracker} \\ \textit{box}} & \tabincell{c}{\textit{scissors}} & \tabincell{c}{\textit{bleach} \\\textit{cleanser}}\\
        \midrule
            Our \textit{Clas} sym & 13.00 & 74.95 & 63.68 & 88.10 & 91.66 \\
            Our \textit{Clas} sym \textbf{+ Arti} & \textbf{12.80} & \textbf{52.70} & \textbf{46.13} & \textbf{66.52} & \textbf{72.31} \\
        \bottomrule
        \end{tabular}
    }
    \end{center}
    \vspace{-5.5mm}
    \caption{Our results ($mm$) on \textbf{DexYCB}. $\bm{\star}$ We only list the MSSD score of 4 objects. The full table can be found in \suppl.}\vspace{-1mm}
    \label{table:dexycb_res}
\end{table}

\begin{table}[t]
    \renewcommand{\arraystretch}{1.0}
    \begin{center}
    \resizebox{1.0\linewidth}{!}{
        \begin{tabular}{l|cc|ccc}
        \toprule
            \multirow{3}{*}{Method} & \multirow{3}{*}{MPJPE} & \multirow{3}{*}{MPCPE}  & \multicolumn{3}{c}{MSSD} \\ \cline{4-6}
            & & & \tabincell{c}{\textit{mustard} \\ \textit{bottle} } & \tabincell{c}{\textit{bleach} \\ \textit{cleanser}} & \tabincell{c}{\textit{potted} \\ \textit{meat can}} \\
        \midrule
            Our \textit{Clas} & 2.94 & 7.53 & 6.88 & 5.56 & 7.63 \\
            Our \textit{Clas} \textbf{+ Arti} & 2.50 & 5.88 & 3.79 & \textbf{4.99} & 6.21 \\
        \midrule
            Our \textit{Clas} sym & 2.98  & - & 3.73 & 6.39 & 7.28 \\
            Our \textit{Clas} sym \textbf{+ Arti} & \textbf{2.34} & - & \textbf{2.66} & 5.23 & \textbf{5.82} \\
        \bottomrule
        \end{tabular}
    }
    \end{center}
    \vspace{-5.5mm}
    \caption{Our results ($cm$) on \textbf{HO3D v3}.}\vspace{-4mm}
    \label{table:ho3dv3_res}
\end{table}

\begin{table*}[ht]
    \begin{center}
        % Complexity
        \begin{minipage}[!htp]{0.33\textwidth}
        \begin{center}
        \makeatletter\def\@captype{table}\makeatother
        \resizebox{\linewidth}{!}{
            \begin{tabular}{l|c|c}
            \toprule
                Training set composition & MPJPE & MPCPE \\
            \midrule
                HO3D & 3.06 & 7.24\\
                \cellcolor{Gray} \textbf{a.1)} HO3D $\bigoplus$ YCBAfford & \cellcolor{Gray} 3.01 & \cellcolor{Gray} 6.89 \\
                \cellcolor{DGray} \textbf{a.2)} HO3D $\bigoplus$ CCV-space & \cellcolor{DGray} 2.71 & \cellcolor{DGray} 5.49 \\
                HO3D \textbf{+ Arti} (full version) & \textbf{2.64} & \textbf{5.16}\\
            \bottomrule
            \end{tabular}
        }
        \end{center}\vspace{-6.5mm}
            \caption{Ablation on grasp poses synthesis. \textbf{a.1) \textit{v.s} a.2)} shows a same network model (\textit{Clas}) trained on HO3D plus synthetic pose from Conventional \textbf{\textit{v.s}} from CCV-space. ($cm$)}
        \label{table:compare_graspit}
        \end{minipage}
        ~~~
        \begin{minipage}[!htp]{0.3\textwidth}
        \begin{center}
            \resizebox{1.0\linewidth}{!}{
            \makeatletter\def\@captype{table}\makeatother
            \setlength{\tabcolsep}{1mm}{
                \begin{tabular}{l|c|c}
                    \toprule
                        Method on \% of source data  & MPJPE & MPCPE \\
                    \midrule
                        Our \textit{Reg} (10\%) & 3.81 & 8.77\\
                        Our \textit{Reg} (100\%) & 3.53 & 7.38\\
                        Our \textit{Reg} (10\%) \textbf{+ Arti} & \textbf{3.29} & \textbf{6.87}\\
                        \midrule
                        Our \textit{Clas} (10\%) & 3.63 & 7.66\\
                        Our \textit{Clas} (100\%) & 3.06 & 7.24\\
                        Our \textit{Clas} (10\%) \textbf{+ Arti} & \textbf{3.05} & \textbf{6.02}\\
                    \bottomrule
                \end{tabular}
            }
        }
        \end{center}\vspace{-5.5mm}
            \caption{Performance of models trained on 10\% of \textbf{HO3D} source data. ($cm$)}
        \label{tab:f01_ho3d_res}
        \end{minipage}
        ~~~
        \begin{minipage}[!htp]{0.31\linewidth}
        \begin{center}
            \resizebox{1.0\linewidth}{!}
            {
            \makeatletter\def\@captype{table}\makeatother
            \setlength{\tabcolsep}{1mm}{
                \begin{tabular}{l|l|cc|c}
                    \toprule
                        Dataset & Method & MPJPE & MPCPE & CS-J \\
                    \midrule
                        HO3Dv1 & \cite{hasson2020leveraging} & 5.75& 9.61 & 6.24 \\
                        HO3Dv1 & \cite{hasson2020leveraging} + \textbf{Arti} & \textbf{3.67} & \textbf{3.24} & \textbf{3.57} \\
                    \midrule
                        HO3D & \cite{hasson2020leveraging}  & 3.69 & 12.38 & 5.52 \\
                        HO3D & \cite{hasson2020leveraging} \textbf{+ Arti}  & \textbf{3.39} & \textbf{8.31} & \textbf{4.90} \\
                    \bottomrule
                \end{tabular}
                }
            }
        \end{center}\vspace{-3.5mm}
            \caption{Results of porting ArtiBoost to the model in Hasson \etal \cite{hasson2020leveraging}. \textbf{CS-J}: the MPJPE in camera space; all in $cm$.} 
            \label{table:expansibility}
    \end{minipage}
    \end{center}\vspace{-7mm}
\end{table*}

\vspace{-0.5mm}\subsection{Ablation Study}\label{sec:ablation}
\vspace{-1 mm}
To further discover how ArtiBoost works, we design two ablation studies. 
\textbf{A).} We compare the ArtiBoost with conventional grasp synthesis methods to show the efficacy of the CCV-space. \textbf{B).} We compare ArtiBoost with an offline training scheme to demonstrate the efficiency of our dynamic online re-weighting.

\vspace{0.5mm}
\noindent\textbf{A). Conventional Grasp Synthesis.} 
Simulated hand-object poses in GraspIt \cite{graspit} are not necessarily correct or diverse (see Sec. 3.1 and Fig. 3 of GanHand \cite{corona2020ganhand}). Therefore,  
Corona \etal \cite{corona2020ganhand} manually annotated the MANO hand grasping YCB objects \cite{ycb} and released a grasping pose dataset: YCBAfford. 
We find YCBAfford is an ideal contrast of our synthetic grasps in CCV-space. 
In this study, we compare the performance of the same model trained on two different data compositions. One is \textbf{a.1)} HO3D plus YCBAfford, and the other is \textbf{a.2)} HO3D plus our synthetic poses in CCV-space.
To ensure fair and instructive comparison, we use the same amounts of HO poses from YCBAfford and CCV-space, set up an identical rendering pipeline, and turn off the re-weighting.
During training, the synthetic HO poses will be randomly sampled and rendered, and then the rendered images will be blended into the original HO3D training set. 
\cref{table:compare_graspit} shows the results of this study. We find that the model trained with poses in CCV-space outperforms the model trained with conventional synthetic poses, verifying an essential idea in our paper: diverse pose variants facilitate pose estimation.

\begin{figure}[t]
    \begin{center}
    \includegraphics[width=1.0\linewidth]{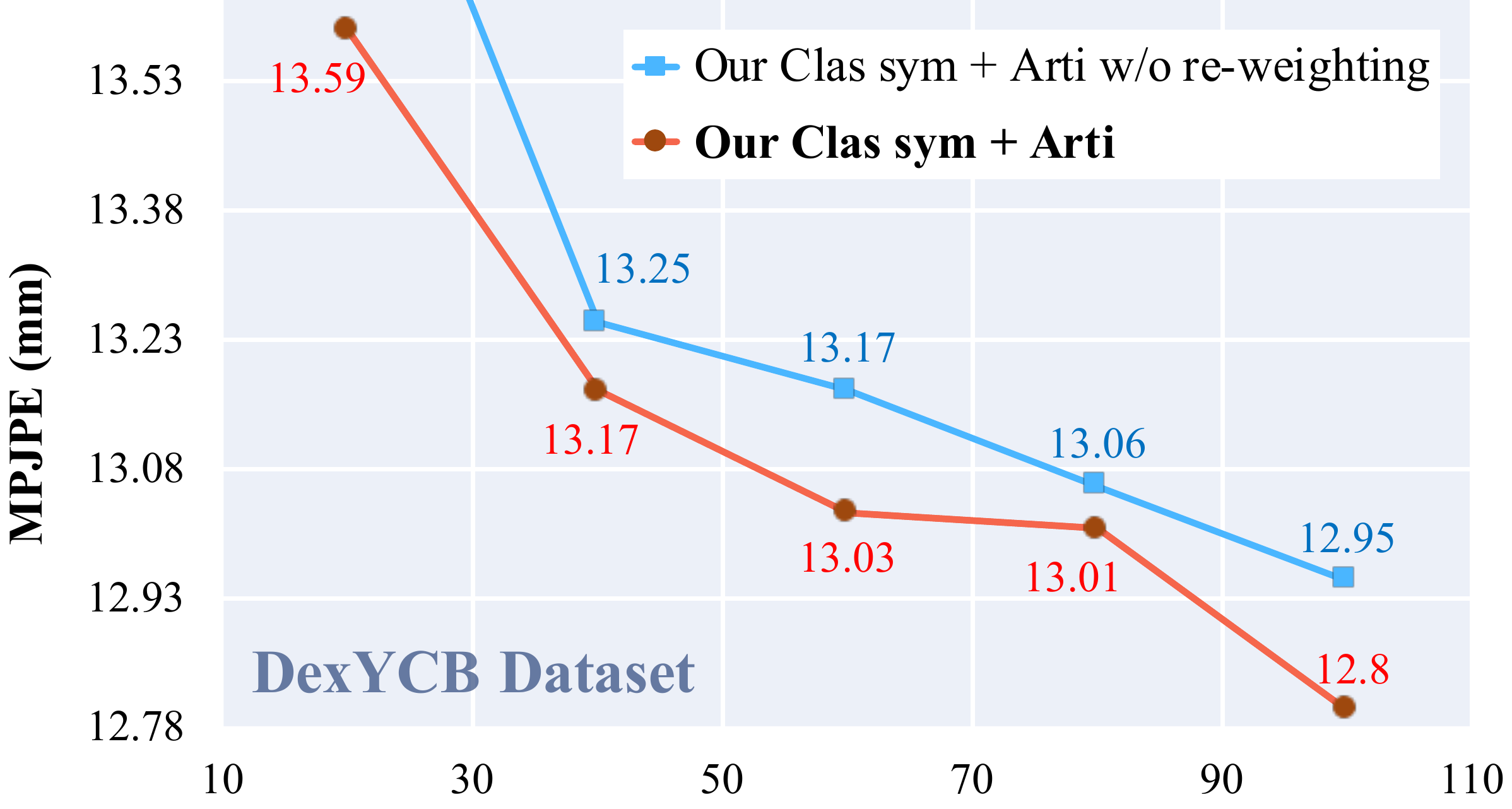}\vspace{-6mm}
    \end{center}
       \caption{Ablation on offline training scheme.}\vspace{-4mm}
    \label{fig:offline_ablation}
\end{figure}

\vspace{0.5mm}
\noindent\textbf{B). Offline Training Scheme.} 
To simulate the offline training scheme, we fix all the weights in the sampling weight map and randomly choose triplets from the CCV-space. 
We inspect two experiments throughout the entire training process, one of which uses the online sample re-weighting strategy, and the other follows the offline scheme. Both experiments use the \textit{Clas} and symmetry model and are trained on the DexYCB dataset.  
We report the interim results on DexYCB testing set at certain intervals. As shown in \cref{fig:offline_ablation}, online sample re-weighting helps the models to converge fast and achieve a higher score.

\vspace{-1 mm}\subsection{Applications}\label{sec:application}
\vspace{-1 mm}
We explore the potential application of ArtiBoost and design two studies.
\textbf{A).} As real-world data labeling is inefficient and costly, 
we try to use ArtiBoost to help neural models on training with less amount of real-world labeled data.
\textbf{B).} As ArtiBoost is model-agnostic, we show that it can be ported to other HOPE learning frameworks and boost their performance.

\vspace{0.5mm}\noindent\textbf{A). Training on Less Real-world Labeled Data.}
~This study trains the neural models using only a small portion of the HO3D training data. 
Supposing the original amount of data in the HO3D training set is $N$, we set up three different amounts of training set: (1) 10\% $N$ of the original set, (2) 100\% $N$ of the original set, and (3) 10\% $N$ of the original set plus 100\% $N$ of ArtiBoost synthetic poses. 
As shown in \cref{tab:f01_ho3d_res}, we find that neural networks trained with setting (3) can outperform the network trained with 100\% real-world data.

\vspace{0.5mm}\noindent\textbf{B). Porting ArtiBoost to other HOPE Model.}
~In \cref{table:expansibility}, we provide the results of porting ArtiBoost to another HOPE framework proposed by Hasson \etal  \cite{hasson2020leveraging} which directly regressed the hand-object poses and focal-normalized camera-space translations.  
The source training dataset: \textbf{HO3Dv1} \cite{hampali2020ho3dv1} used in \cite{hasson2020leveraging} is an early version of HO3D. 
We reproduced the results in \cite{hasson2020leveraging} by training their network on the predefined \textbf{v1} set only. We report both the MPJPE in camera space and wrist-relative system. 
We show in \cref{table:expansibility} that porting ArtiBoost into a camera-space HOPE model significantly improves all metrics.

\vspace{-2mm}\section{Discussion}
\vspace{-1mm}\noindent\textbf{Limitation.} 
However, we do not explicitly mitigate the domain gap between the synthetic and real data, as we find that the dominant improvement to the HOPE task is brought from the images of more diverse pose variants, rather than images with a more realistic appearance. 
Besides, as the renderer in ArtiBoost is not differentiable, current ArtiBoost only supports exploration in a predefined lookup table (\eg CCV-space). In future work, we will investigate a powerful generative and contrastive model seeking common features shared by both real and synthetic images. 

\vspace{1mm}\noindent\textbf{Conclusion.} In this work, we propose a novel online data enrichment method ArtiBoost, which enhances the learning framework of articulated pose estimation by exploration and synthesis. 
Our proposed ArtiBoost can be integrated into any learning framework, and in this work, we show its efficacy on the challenging task of hand-object pose estimation. Even with a simple baseline, our method can boost it to outperform the previous SOTA on the popular datasets.
Besides, the proposed CCV-space also opens the door towards the generic articulated pose estimation, which we leave as future work.

\noindent\rule{\columnwidth}{1.5pt}

\noindent\textbf{Acknowledgment} This work was supported by the National Key Research and Development Project of China (No.2021ZD0110700), Shanghai Municipal Science and Technology Major Project (2021SHZDZX0102), Shanghai Qi Zhi Institute, and SHEITC (2018-RGZN-02046).

{
\small
\balance
\bibliographystyle{config/ieee_fullname}
\bibliography{egbib.bib}
}

\clearpage
\begin{appendices}
\label{appendices}

\vspace{-1 mm}\section{The Training Details}
\vspace{-1 mm}
The backbones of classification-based (\textit{Clas}) and regression-based (\textit{Reg}) baseline networks are initialized with ImageNet \cite{deng2009imagenet} pretrained model. 
In \textit{Clas}, the output resolution of 3D-heatmaps is 28 $\times$ 28 $\times$ 28. The MLP branch that predicts object rotation adopts three fully-connected layers with 512, 256 and 128 neurons for each, and a final layer of 6 neurons that predict the continuity representation \cite{zhou2019continuity} of object rotation: $\mathbf{r}_o \in \mathfrak{so}(3)$.
We train the network 100 epochs with Adam optimizer and learning rate of $ 5\times 10^{-5}$. 
The training batch size across all the following experiments is 64 per GPU and 2 GPUs in total.
The framework is implemented in PyTorch. All the object models and textures are provided by the original dataset.
For all the training batches, the blended rate of original real-world data and ArtiBoost synthetic data is approximately $1:1$. We empirically find that this real-synthetic blended rate achieves the best performance.

\section{Objects' Symmetry Axes}\label{sec:symmetry_axes}

In the hand-object interaction dataset, it is far more challenging to predict the pose of an object than in the dataset that only contains objects, since the objects are often severely occluded by the hand. Therefore, we relax the restrictions of the objects' symmetry axes following the practices in \cite{chao2021dexycb,hampali2021handsformer}. 
Supposing the set $\mathcal{S}$ contains all the valid rotation matrices based on the object's predefined symmetry axes,
we calculate $\mathcal{S}$ with the following step:

\begin{table}[h]
  \renewcommand{\arraystretch}{1.0}
  \begin{center}
  \resizebox{1.0\linewidth}{!}{
      \begin{tabular}{c|c|c}
          \toprule
              Objects  & Axes: $\mathbf{n}$ & Angle: $\theta$ \\
              \midrule
              002\_master\_chef\_can & x, y, z & $180^\circ$, $180^\circ$, $\infty$\\
              003\_cracker\_box & x, y, z & $180^\circ$, $180^\circ$, $180^\circ$\\
              004\_sugar\_box & x, y, z & $180^\circ$, $180^\circ$, $180^\circ$\\
              005\_tomato\_soup\_can & x, y, z & $180^\circ$, $180^\circ$, $\infty$\\
              006\_mustard\_bottle & z & $180^\circ$\\
              007\_tuna\_fish\_can & x, y, z & $180^\circ$, $180^\circ$, $\infty$\\
              008\_pudding\_box & x, y, z & $180^\circ$, $180^\circ$, $180^\circ$ \\
              009\_gelatin\_box & x, y, z & $180^\circ$, $180^\circ$, $180^\circ$ \\
              010\_potted\_meat\_can & x, y, z & $180^\circ$, $180^\circ$, $180^\circ$ \\
              024\_bowl & z & $\infty$\\
              036\_wood\_block & x, y, z & $180^\circ$, $180^\circ$, $90^\circ$ \\
              037\_scissors & z & $180^\circ$\\
              040\_large\_marker & x, y, z & $180^\circ$, $\infty$, $180^\circ$\\
              052\_extra\_large\_clamp & x & $180^\circ$\\
              061\_foam\_brick & x, y, z & $180^\circ$, $90^\circ$, $180^\circ$ \\
          \bottomrule
  \end{tabular}
  }
  \end{center}\vspace{-5mm}
  \caption{\textbf{YCB objects' axes of symmetry}. $\infty$ indicates the object is revolutionary by the axis.}
  \label{tab:def_sym}
\end{table}

\begin{figure}[h]
  \begin{center}
    \includegraphics[width=1.0\linewidth]{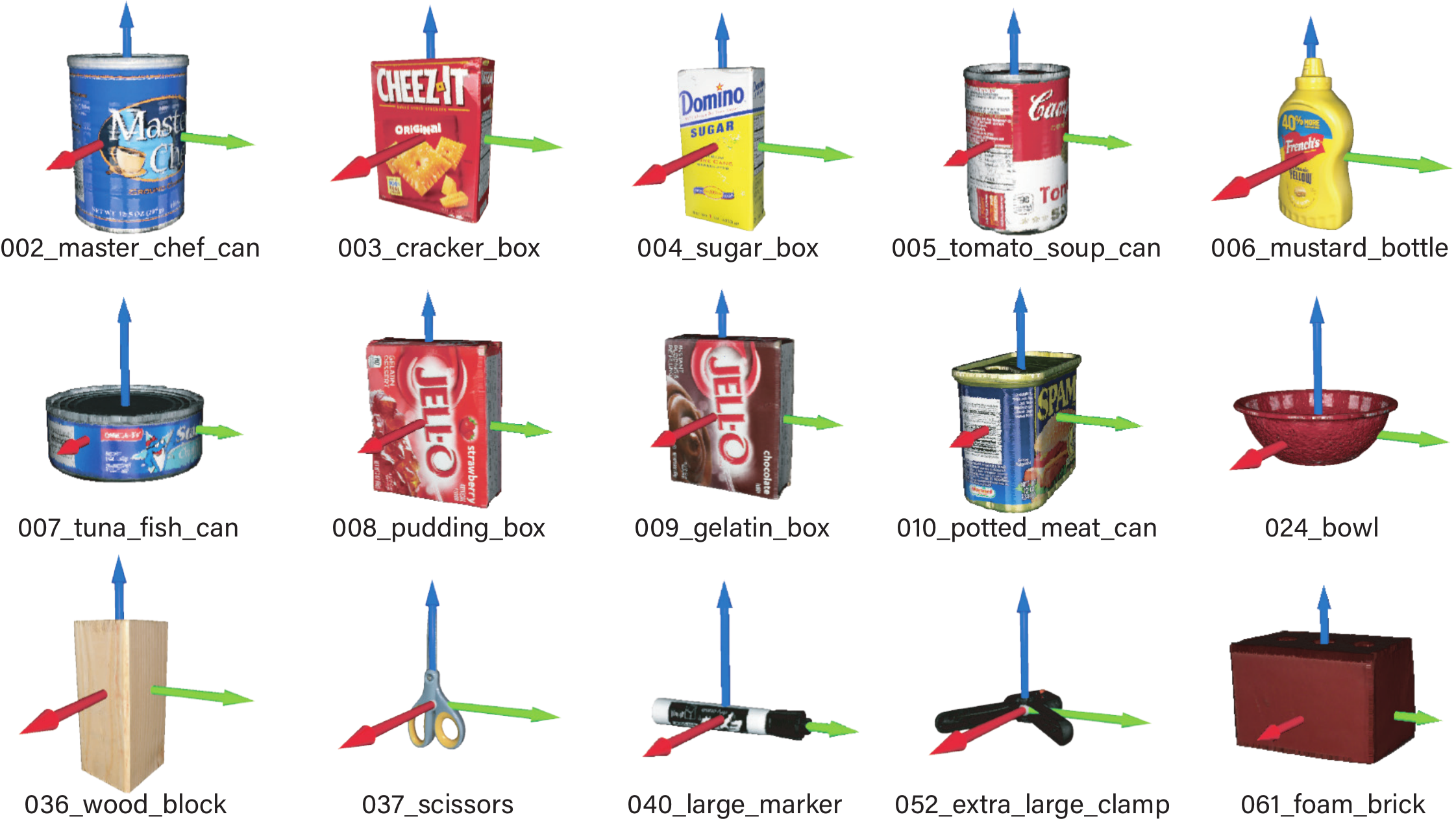}\vspace{-3mm}
  \end{center}
    \caption{\textbf{YCB objects' principal axis of inertia.} The x, y and z axis are colored in red, green and blue, respectively.}\vspace{-3mm}
  \label{fig:pca_obj}
\end{figure}

\begin{enumerate}[label={\arabic*)},font={\bfseries},leftmargin=*]
    \setlength\itemsep{-1 mm}
    \item Firstly, as shown in Fig~\ref{fig:pca_obj}, we align the object to its principal axis of inertia.
    \item Secondly, we define the axis $\mathbf{n}$ and angle $\theta$ of symmetry in Tab~\ref{tab:def_sym} under the aligned coordinate system, where the object's geometry does not change when rotate this object by an angle of $\theta$ around $\mathbf{n}$. Here we get the predefined rotation matrix $\mathbf{R}_{def} = \exp(\theta\mathbf{n})$.
    \item To get a more accurate rotation matrix $\mathbf{R}$, we use the Iterative Closest Point (ICP) algorithm to fit a $\Delta \mathbf{R}$. The ICP minimizes the difference between $\Delta \mathbf{R} * \mathbf{R}_{def} * \mathbf{V}_o$ and $\mathbf{V}_o$, where $\mathbf{V}_o$ is the point clouds on object surface. Finally, we have $\mathbf{R} = \Delta \mathbf{R} * \mathbf{R}_{def}, \mathbf{R} \in \mathcal{S}$.
\end{enumerate}

\begin{table*}[ht]
  \renewcommand{\arraystretch}{1.0}
  \begin{center}
  \resizebox{0.95\linewidth}{!}{
      \begin{tabular}{c|c|c|c|c|c}
          \toprule
              Objects & Our \textit{Clas} sym &  Our \textit{Clas} sym \textbf{+ Arti} & Objects & Our \textit{Clas} sym &  Our \textit{Clas} sym \textbf{+ Arti}\\
              \midrule
              002\_master\_chef\_can & 27.62 &  \textbf{25.59}     & 003\_cracker\_box & 63.68 & \textbf{46.13}\\
              004\_sugar\_box & 48.42 & \textbf{39.20}             & 005\_tomato\_soup\_can & 33.31 & \textbf{31.90}\\
              006\_mustard\_bottle & 35.16 & \textbf{32.01}        & 007\_tuna\_fish\_can & 24.54 & \textbf{23.81}\\
              008\_pudding\_box & 39.92 & \textbf{35.04}           & 009\_gelatin\_box & 45.99 & \textbf{37.81}\\
              010\_potted\_meat\_can & 41.44 & \textbf{36.47}      & 011\_banana & 98.69 & \textbf{79.87}\\
              019\_pitcher\_base & 105.66 & \textbf{84.82}         & 021\_bleach\_cleanser & 91.66 & \textbf{72.31}\\
              024\_bowl & \textbf{31.74} & 32.37                   & 025\_mug  & 65.46 & \textbf{54.28}\\
              035\_power\_drill & 74.95 & \textbf{52.70}           & 036\_wood\_block & 51.24 & \textbf{50.69}\\
              037\_scissors & 88.10 & \textbf{66.52}               & 040\_large\_marker & 30.76 & \textbf{29.33}\\
              052\_extra\_large\_clamp & 78.87 & \textbf{55.87}    & 061\_foam\_brick & 34.23 & \textbf{31.53}\\
          \bottomrule
  \end{tabular}
  }
  \end{center}
  \vspace{-5mm}
  \caption{Full MSSD results ($mm$) on \textbf{DexYCB} testing set.}\vspace{-1mm}
  \label{tab:full_dexycb}
\end{table*}

\begin{figure*}[!t]
  \begin{center}
    \includegraphics[width=1.0\linewidth]{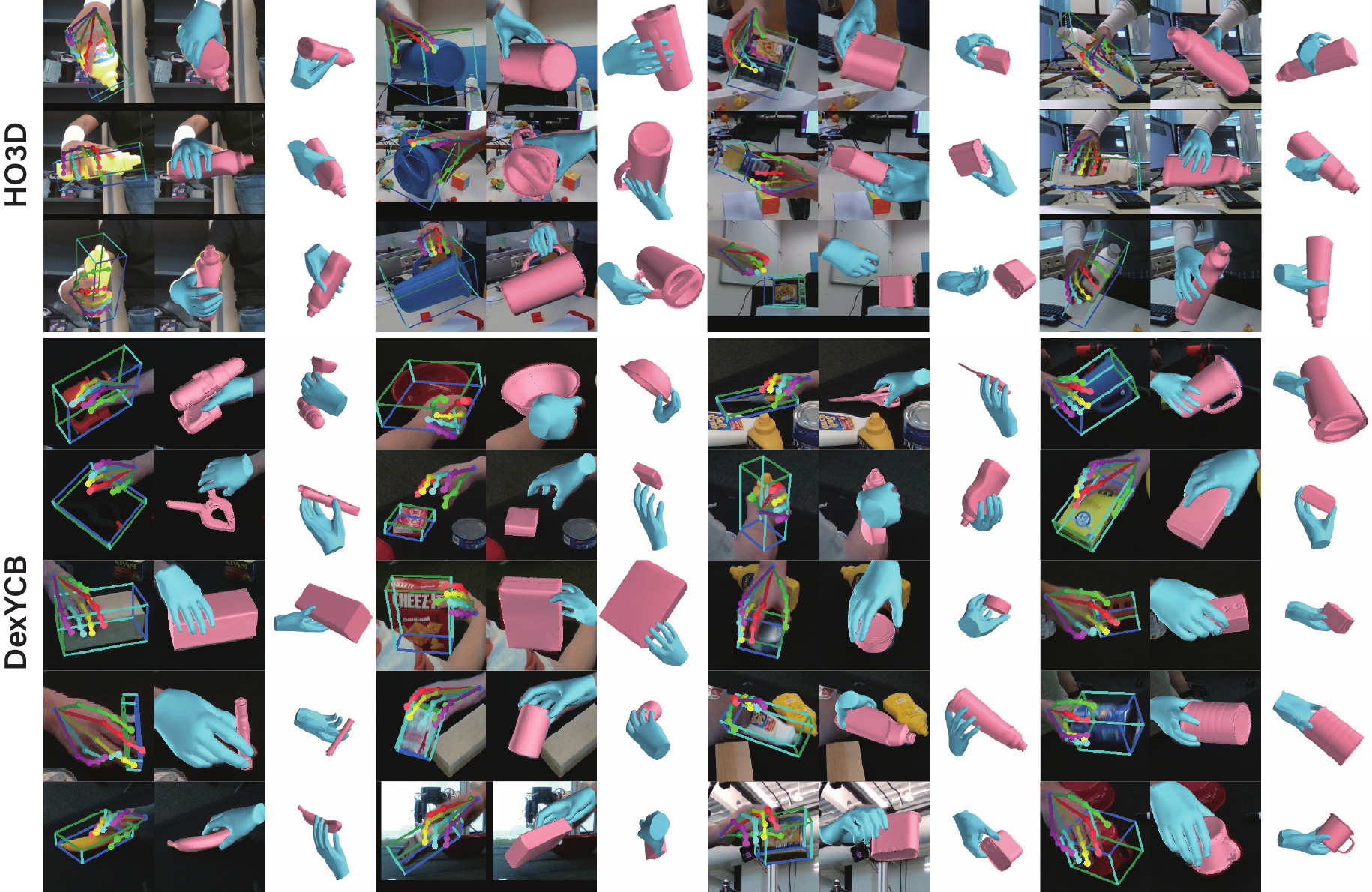}
  \end{center}
  \vspace{-5mm}
    \caption{(\textbf{Best view in color}) More qualitative results on \textbf{HO3D} (\engordnumber{1} $\sim$ \engordnumber{3} rows) and \textbf{DexYCB} (\engordnumber{4} $\sim$ \engordnumber{8} rows) datasets.}\vspace{-1mm}
  \label{fig:supp_qua}
\end{figure*}

\section{Additional Results}\label{sec:add_results}

We demonstrate 20 YCB objects' MSSD on DexYCB in \cref{tab:full_dexycb}. With ArtiBoost, our network can predict a more accurate pose for almost every object.
More qualitative results on HO3D and DexYCB testing set are shown in \cref{fig:supp_qua}.

\end{appendices}

\end{document}